\documentclass{article} %
\usepackage[T1]{fontenc}
\usepackage{iclr2027_conference,times}
\iclrfinalcopy

\usepackage{amsmath,amssymb,bm}
\usepackage{graphicx}
\usepackage{booktabs,array,multirow,threeparttable}
\usepackage{xcolor}
\usepackage{xspace}
\usepackage{caption}
\usepackage{subcaption}
\usepackage[shortlabels,inline]{enumitem}
\usepackage{tikz}
\usetikzlibrary{arrows.meta,positioning,calc,fit,backgrounds,shapes.geometric}
\usepackage{microtype}
\usepackage{hyperref}
\hypersetup{hidelinks}
\usepackage{url}
\usepackage[capitalize]{cleveref}
\crefname{section}{Sec.}{Secs.}\Crefname{section}{Section}{Sections}
\crefname{table}{Tab.}{Tabs.}\Crefname{table}{Table}{Tables}
\crefname{appendix}{Appendix}{Appendices}\Crefname{appendix}{Appendix}{Appendices}
\crefname{equation}{Eq.}{Eqs.}\Crefname{equation}{Equation}{Equations}
\graphicspath{{figs/}}

\makeatletter\DeclareRobustCommand\onedot{\futurelet\@let@token\@onedot}
\def\@onedot{\ifx\@let@token.\else.\null\fi\xspace}\makeatother

\newcommand{\papertitle}{Dense4D:\\Pixel-Level 4D Human Correspondence}
\newcommand{\method}{Ours\xspace}
\newcommand{\R}{\mathbb{R}}
\newcommand{\bx}{\mathbf{x}}

\newcommand{\bX}{\mathbf{X}}
\newcommand{\bK}{\mathbf{K}}
\newcommand{\bR}{\mathbf{R}}
\newcommand{\bt}{\mathbf{t}}
\newcommand{\bI}{\mathbf{I}}
\DeclareMathOperator*{\argmin}{arg\,min}

\newcommand{\best}[1]{\textbf{#1}}

\title{\papertitle}

\author{R{\i}za Alp G{\"u}ler\\
\normalfont Department of Computing, Imperial College London}

\begin{document}

\maketitle
\fancyhead{}
\renewcommand{\headrulewidth}{0pt}
\suppressfloats[t]

\begin{abstract}
\urlstyle{same}
Reconstructing human motion in 4D requires identifying the same surface points through articulation, deformation and changes in visibility.
We introduce Dense4D, to our knowledge the first feed-forward model dedicated to dense 4D correspondence across the body, clothing and hair from a set of monocular RGB images, without a body template.
Trained specifically for human geometry, Dense4D jointly processes person-centred crops with known image-to-crop transforms to predict Human Point Maps (HPMs) and depth for each image.
An HPM assigns each person pixel the 3D position of its observed surface point at a common reference time.
These maps support amodal correspondence: points hidden in the reference image can be localised from observations in other frames.
Matching HPM coordinates across crops establishes correspondence without predicting camera poses or large scene displacements; per-image depth then lifts the matches into 3D tracks.
Experiments on real people demonstrate substantially higher tracking accuracy and closer agreement with scanned surfaces than the evaluated baselines.
Points from different frames assemble into coherent human geometry in the reference pose, providing visible evidence of consistent dense correspondence through articulation and clothing deformation.
Additional qualitative results, videos, and an interactive viewer are available at \href{https://alpguler.com/dense4d}{\textcolor{blue!55!black}{\textbf{https://alpguler.com/dense4d}}}.

\end{abstract}

\noindent\begin{minipage}{\textwidth}
\centering
\begin{tikzpicture}
\node[inner sep=0,outer sep=0,anchor=south west] (art) at (0,0) {\includegraphics[width=\textwidth]{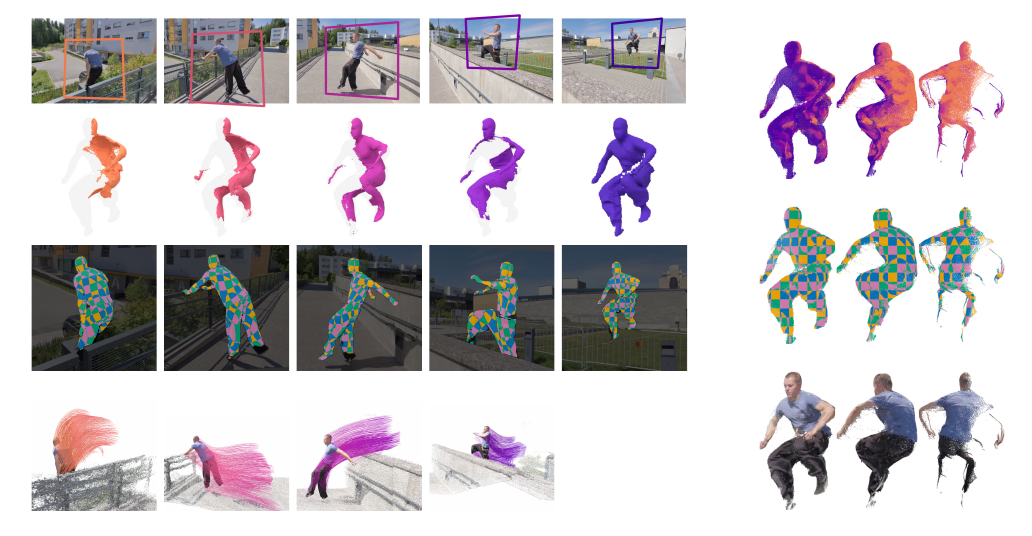}};
\begin{scope}[x={(art.south east)},y={(art.north west)}]
\tikzset{stage/.style={rotate=90,anchor=base west,inner sep=0,outer sep=0,font=\fontsize{6.5pt}{7.5pt}\selectfont}}
\node[stage] at (0.0203,0.856) {Input};
\node[stage] at (0.009,0.623) {Per-frame};
\node[stage] at (0.027,0.645) {HPMs};
\node[stage] at (0.0142,0.388) {Dense};
\node[stage] at (0.0263,0.326) {Correspondence};
\node[stage] at (0.0203,0.042) {Dense 4D Tracking};
\node[anchor=base,inner sep=0,outer sep=0,font=\fontsize{7.5pt}{8.5pt}\selectfont] at (0.858,0.959) {Human Point Map (HPM) union};
\end{scope}
\end{tikzpicture}

\captionsetup{font=small,skip=3pt}
\captionof{figure}{\textbf{Dense 4D correspondence of the body, clothing and hair.}
Top left: boxes identify input frames; each HPM below maps visible person pixels into the reference pose, coloured by source frame.
Right: the HPM union from three viewpoints, coloured by source frame (top), a reference-space 3D chequerboard (middle), and RGB (bottom).
The alignment of points across frames reveals a coherent body despite motion.
The chequerboard transferred to the images (third row, left) visualises dense correspondence.
Matching HPM coordinates and lifting target pixels with predicted depth produces the 3D tracks at bottom left.}
\label{fig:teaser}
\end{minipage}

\section{Introduction}\label{sec:intro}

A detailed understanding of human geometry and motion is important across many fields, including robotics, extended reality, entertainment, healthcare, sports analytics, and sign language understanding.
Yet people remain challenging for general dynamic scene reconstruction systems because their motion combines body articulation with the independent deformation of clothing and hair.
Following these surfaces through a video requires identifying the same points as they move, become occluded and reappear.
Reconstructing each frame independently does not establish this correspondence.
We study \emph{dense 4D human correspondence}, the task of matching points across the body, clothing and hair in different images and recovering their three-dimensional trajectories.

Human correspondence methods commonly map image pixels to a shared body template~\citep{guler2018densepose}.
Such coordinates describe the underlying body, but do not directly represent the geometry and independent motion of loose garments and hair.
In parallel, many 4D dynamic scene reconstruction systems jointly estimate geometry, motion and cameras across diverse scenes~\citep{sucar2026vdpm,zhang2026d4rt}.
We introduce Dense4D for dense 4D human correspondence without a body template.
The model establishes correspondence on the person without jointly reconstructing the surrounding scene or predicting camera poses.

\begin{figure}[t]
\centering
\input{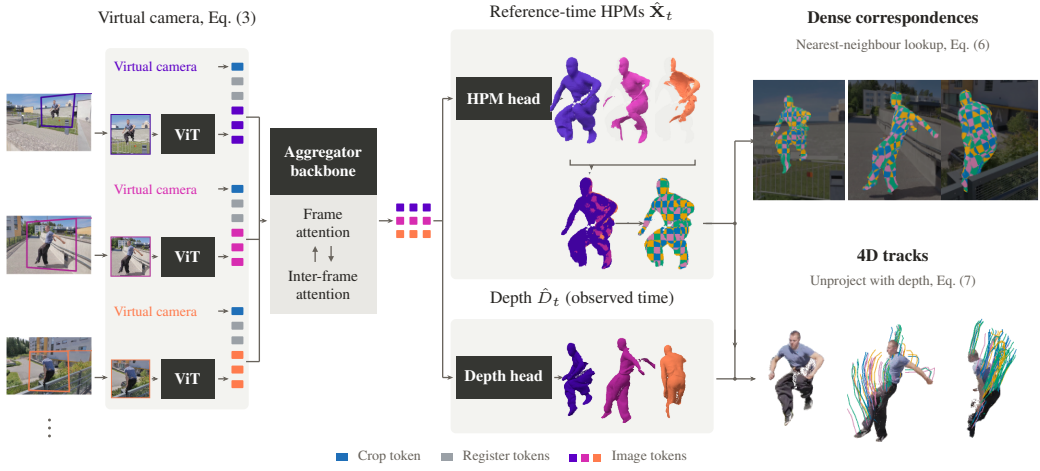}
\caption{\textbf{Dense 4D human correspondence.}
An extra crop token conditions the aggregator on cropping geometry (\cref{eq:cond}).
The heads predict reference-time HPMs and per-image depth, visualised as HPM point clouds and unprojected depth maps.
Nearest-neighbour lookup matches HPM coordinates within the target frame (\cref{eq:nn}).
A 3D chequerboard defined in reference space colours the union and is transferred to the images through HPM coordinates, visualising dense correspondence.
Unprojecting the matched pixels with predicted depth gives 4D tracks (\cref{eq:owntime,sec:method_inference}), shown as coloured trajectories.}
\label{fig:method}
\end{figure}

Dense4D follows a top-down approach: we isolate each person in the scene and jointly process their crops across time (\cref{fig:method}).
Using the focal length and person mask, a known virtual-camera warp reorients each crop towards the person, and the crop zoom is supplied to the network.
A Human Point Map (HPM) assigns each person pixel the 3D position of its observed surface point at a selected reference time, expressed in the reference crop's camera coordinates.
These maps support \emph{amodal correspondence} to the reference surface, including regions hidden in the reference image but observed in other frames.
Their union samples this reference-time geometry (\cref{fig:teaser}, right).

Where methods such as V-DPM predict point motion in a shared camera frame~\citep{sucar2026vdpm}, Dense4D identifies corresponding pixels by matching HPM coordinates across person-centred crops.
The learned mapping relates observations of the person with known crop transforms and zoom, without also predicting the person's displacement through the scene.
Each matched pixel is lifted with the depth predicted for its image to recover the 3D track position.
External scene geometry and camera poses are used separately to place the reconstructed person and tracks in world coordinates.

We use the image encoder and aggregator architecture of \citet{wang2026vggtomega}.
An additional crop token conditions the aggregator, which alternates attention within and across frames.
Two dense heads predict HPMs and per-image depth in a single forward pass.
Training uses human geometry and correspondence supervision from synthetic motion and multi-view renderings of real scans.

Across articulated motion and clothing deformation, Dense4D brings observations from different frames into coherent reference-time human geometry, with visibly aligned body parts and fine surface details (\cref{sec:exp_wild}).
On real footage, Dense4D recovers more accurate human trajectories and better matches the scanned reference geometry than the methods we compare against (\cref{sec:exp_pstudio,sec:exp_4ddress}).
The shared reference representation also supports tracking across long intervals, as demonstrated on an 86\,s video (\cref{fig:long_row}).

Our contributions are:
\begin{itemize}[leftmargin=1.4em,itemsep=2pt,topsep=3pt,parsep=0pt]
  \item We introduce Dense4D, to our knowledge the first feed-forward model dedicated to template-free dense 4D human correspondence from a set of monocular RGB images.
Its predictions assemble observations of the body, clothing and hair into a coherent reference-time geometry, including regions hidden in the reference image.
  \item We decouple correspondence estimation from camera-pose and scene-motion recovery.
Dense4D supports direct matching across long temporal intervals and integrates with scene reconstruction through shared image pixels.
  \item We introduce an amodal 3D localisation protocol on 4D-DRESS, measuring how closely observations across time align with the full reference-time scan.
We also evaluate 4D point tracking on the human queries of the existing TAPVid-3D Panoptic Studio benchmark.
\end{itemize}

\newpage
\section{Related work}\label{sec:related}

\paragraph{Human correspondence and parametric reconstruction.}
A shared body template allows pixels on different people to be related through common anatomical coordinates~\citep{guler2018densepose}.
BodyMap and TemporalUV extend this association to clothing and hair~\citep{ianina2022bodymap,zhang2022temporaluv}, while HumanGPS learns scan-supervised embeddings shared across subjects~\citep{tan2021humangps}.
These coordinates and embeddings establish correspondence but do not themselves reconstruct the individual's body, garment or hair geometry.
Mesh reconstruction methods such as 4DHumans~\citep{goel2023hmr2} achieve body correspondence thanks to the fixed topology of the underlying parametric model, but do not represent the geometry and independent motion of garments or hair.
Dense4D provides \emph{personalised correspondence}: HPMs reconstruct reference-time points and locate later observations on them, without a shared body template.

\paragraph{Deformable human reconstruction.}
ReLoo and DressRecon fit personalised surfaces, appearance and body/clothing deformation models to video through differentiable rendering~\citep{guo2025reloo,tan2025dressrecon}.
Their temporal correspondences follow the fitted 3D deformation fields used to render the reconstructed motion.
Forge4D reconstructs pixel-aligned 3D Gaussians from uncalibrated sparse-view videos and predicts their motion between adjacent timestamps for novel-view and novel-time synthesis~\citep{hu2026forge4d}.
Dense4D establishes correspondence across a set of monocular RGB images through a common human reference-time geometry, with correspondence and depth learned from dense 3D supervision.

\paragraph{Point tracking and amodal correspondence.}
From Particle Video~\citep{sand2006particlevideo} to learned models~\citep{harley2022pips,doersch2023tapir,karaev2025cotracker3}, image point trackers estimate query trajectories through occlusion.
Three-dimensional trackers add depth or geometric estimation~\citep{xiao2024spatialtracker,xiao2025spatialtrackerv2,ngo2025delta,zhang2025tapip3d}.
HPMs also locate points in 3D at the reference time even when hidden in the reference image.

\paragraph{Object-centric representations.}
NOCS predicts category-level object coordinates~\citep{wang2019nocs}; CaSPR learns a canonical spatiotemporal space from dynamic point clouds~\citep{rempe2020caspr}.
DualPM pairs posed and rest-pose maps for quadrupeds~\citep{kaye2025dualpm}; Face Anything predicts template-aligned facial coordinates and depth~\citep{kocasari2026faceanything}.
Dense4D instead matches observations of a deforming person using that person's own reference-time geometry.

\paragraph{Scene reconstruction.}
Point-map models represent pixels from different images as 3D points in a shared reference camera frame~\citep{wang2024dust3r,leroy2024mast3r,wang2025vggt}.
Human-aware models add person outputs: HAMSt3R adds segmentation and DensePose to MASt3R~\citep{rojas2025hamst3r}, while Human3R recovers SMPL-X bodies with scene geometry and cameras from monocular video~\citep{chen2026human3r}.

Feed-forward models extend point-based reconstruction to dynamic scenes~\citep{feng2025st4rtrack,zhang2026d4rt,jiang2026omnix}, often jointly training geometry, motion and camera predictors~\citep{karhade2025any4d,luo2026fourrc,sucar2026vdpm}.
Building on predicted geometry, 4D Primitive-M{\^a}ch{\'e} uses dense image correspondences to optimise rigid-primitive motion and aggregate observations across time~\citep{mazur2026fourdpm}.
Dynamic Point Maps and V-DPM are closest to our representation: they express observations at a shared reference time and viewpoint, allowing correspondence to be established by matching shared-space coordinates~\citep{sucar2025dpm,sucar2026vdpm}.
Dense4D tracks by matching rather than by decoding point motion: every observed person pixel is mapped to the reference surface, correspondences are found among the target frame's observed pixels, and each match is lifted with that frame's own predicted depth.
Tracks therefore lie on the target frame's predicted geometry by construction; visibility is estimated from the matching residual, and no motion or camera-pose prediction is required for camera-space tracking.

\newpage

\section{Method}\label{sec:method}

\subsection{Dense 4D correspondence with Human Point Maps}\label{sec:method_canonical}

Given images $\{\bI_t\}_{t=1}^{S}$ of one person, person masks and focal lengths $f_t$, we predict the geometry and correspondence of the person's body, clothing and hair across time.
Using each mask and focal length, we resample the person into a virtual camera aimed at the centre of the person's bounding box, keeping the camera centre fixed.
At inference, focal length is Dense4D's only supplied camera parameter; the model neither receives nor predicts world-camera poses.
Below, pixel coordinates and foreground sets $\mathcal M_t$ refer to these crops. Ground-truth masks define the training targets; at inference, the masks specify the person to be processed.

The first input is the reference, $r=1$; input order need not be chronological.
For $p\in\mathcal M_t$, the HPM target $\bX_t(p)\in\R^3$ is the position at the reference time of the surface point observed at $p$ in frame $t$.
This position is expressed in the reference image's virtual camera coordinates.
The scalar depth target $D_t(p)$ is the same point's optical-axis depth in frame $t$'s virtual camera at time $t$.
Both targets use a shared normalising scale $s$.

To define these targets formally, let $\bx_{t\to u}(p)$ denote the world-space position at time $u$ of the surface point observed at pixel $p$ in image $t$.
For supervision, the known virtual-camera extrinsics $\mathbf E_{v,t}\in\R^{3\times4}$ map homogeneous world points into the crop's camera coordinates.
The HPM and depth targets are therefore
\begin{equation}\label{eq:target}
\bX_t(p)=\frac{1}{s}\mathbf E_{v,r}
\begin{bmatrix}\bx_{t\to r}(p)\\1\end{bmatrix},
\qquad
D_t(p)=\frac{1}{s}\begin{bmatrix}0&0&1\end{bmatrix}\mathbf E_{v,t}
\begin{bmatrix}\bx_{t\to t}(p)\\1\end{bmatrix}.
\end{equation}
The row vector $[0\;0\;1]$ selects the optical-axis coordinate of the point in the virtual camera.
For the training targets, $s$ is the mean distance of the reference-time surface samples from the reference virtual camera centre:
\begin{equation}\label{eq:normalisation}
s=\frac{1}{\sum_t|\mathcal M_t|}\sum_t\sum_{p\in\mathcal M_t}
\left\|\mathbf E_{v,r}\begin{bmatrix}\bx_{t\to r}(p)\\1\end{bmatrix}\right\|_2.
\end{equation}
All HPMs share the reference virtual camera's origin and axes, and both outputs use the same normalised scale.
The network predicts in this normalised space; absolute body size and camera distance are not recovered.

Observations of the same surface point have equal HPM targets, providing a shared coordinate for correspondence across time.
The union of these targets samples the person's surface at the reference time, bringing together the body, clothing and hair seen across the input clip.
It includes regions hidden in the reference image but observed elsewhere, without fitting a body or surface template.
The network learns to reproduce this common reference surface.

\subsection{Crop-aware prediction}\label{sec:method_inputs}\label{sec:method_network}

We construct the image intrinsics $\bK_t$ from $f_t$, assuming square pixels, zero skew and an image-centred principal point.
The virtual crop is square with side $\ell=512$, intrinsics $\bK_{v,t}$ and a crop-centred principal point.
The rotation $\bR_{v,t}$ maps the ray through the person's box centre from the original camera coordinates to the virtual camera's optical axis.
With a fixed camera centre, the original and virtual image planes are related by a rotation-induced planar homography independent of scene depth~\citep{yu2021pcls}:
\begin{equation}\label{eq:cond}
\mathbf H_t=\bK_{v,t}\bR_{v,t}\bK_t^{-1},\qquad
\mathbf c_t=\big[(\bR_{v,t})_{3,:},\log(f_{v,t}/\ell)\big]\in\R^4.
\end{equation}
Here $f_{v,t}=f_t\ell/\max(c_w,c_h)$, where $c_w$ and $c_h$ are the source-image box width and height.
The first three entries encode the unit box-centre ray in the original camera coordinates; the fourth is the logarithm of the normalised virtual focal length.
Masks define the crops and restrict matching to person pixels.
The network receives the RGB crops and an embedding of these four parameters.
CLIFF and CameraHMR~\citep{li2022cliff,patel2025camerahmr} also supply focal-normalised crop position and size for human mesh recovery.
The \emph{crop embedding} supplies this geometry to the network: an MLP maps $\mathbf c_t$ to one additional token per image (\cref{fig:method}).
This token joins the patch tokens from a ViT image encoder before the first attention block.
The aggregator alternates attention within and across frames to process all observations jointly.
Two dense heads read the aggregator features: the HPM head predicts reference-time surface coordinates $\hat{\bX}_t$, and the depth head predicts depth $\hat D_t$ for each input image. Each head also predicts a confidence map.
The reported model receives all four crop parameters. As expected after homography rectification, fixing the centre-ray inputs to $(0,0,1)$ while retaining crop zoom leaves reconstruction and tracking unchanged at reported precision (Appendix~\ref{sec:exp_cond}).

We initialise the ViT image encoder and aggregator with pretrained VGGT-$\Omega$ weights~\citep{wang2026vggtomega}.
The HPM decoder uses weights from VGGT-$\Omega$'s dense prediction head. Its final layer is initialised from scratch to predict 3D coordinates.
The depth head's final layer is initialised from scratch, and the crop embedding MLP's final layer is initialised to zero.
The model has approximately 974\,M parameters; further architecture and crop-implementation details are given in \cref{sec:supp_method}.

\subsection{Surface supervision and training}\label{sec:method_gt}\label{sec:method_training}

We train with animated surfaces from BEDLAM2.0 (BEDLAM2; \citealp{tesch2025bedlam2}), multi-view renderings of static THuman2.1 scans~\citep{yu2021function4d}, and MammaSyn body-surface tracks~\citep{cuevas2025mamma} (\cref{sec:supp_training}).
\Cref{fig:training_data} in the appendix illustrates the supervision surfaces and input crops.
All three sources provide meshes with consistent vertex and face indices across the input images.
For BEDLAM2, these include body and garment meshes, together with envelope meshes derived from hair grooms and transported with the posed head.
We store each supervised pixel's rasterised triangle index and barycentric coordinates, as also done in Syn4D~\citep{jiang2026syn4d}.
Applying the corresponding perspective-correct weights $\omega_{t,p,m}$ to the reference vertices gives the reference-time position:
\begin{equation}\label{eq:transport}
\bx_{t\to r}(p)=\sum_{m\in F[\varphi_t(p)]}\omega_{t,p,m}\mathbf V_r[m].
\end{equation}
Here $\varphi_t(p)$ is the triangle index, $F[\varphi_t(p)]$ lists its three vertex indices, and $\mathbf V_r[m]$ is vertex $m$ in world coordinates at the reference time.
The same barycentric interpolation~\citep{laine2020nvdiffrast,cole2021rts}, evaluated on frame $t$'s vertices, gives $\bx_{t\to t}(p)$ for the depth target in \cref{eq:target}.
Dataset preparation is detailed in \cref{sec:supp_data}.

For supervised pixels $i=(t,p)$ with $p\in\mathcal M_t$, we minimise the point loss $\mathcal L_{\rm pt}$ and depth loss $\mathcal L_{\rm dp}$ with equal weight.
For each head $k\in\{\mathrm{pt},\mathrm{dp}\}$, let $\sigma_i^k$ be its predicted confidence:
\begin{equation}\label{eq:loss}
\begin{gathered}
\mathcal L_k=\mathcal Q_{.98}\!\left(\{r_i^k\sigma_i^k-0.2\log\sigma_i^k\}_i\right)
+\mathcal Q_{.98}\!\left(\{r_i^k\}_i\right),\\
r_i^{\rm pt}=\|\hat{\bX}_i-\bX_i\|_2,\qquad
r_i^{\rm dp}=|\hat D_i-D_i|,\qquad \sigma_i^k\geq1 .
\end{gathered}
\end{equation}
This confidence-weighted objective follows DUSt3R and VGGT~\citep{wang2024dust3r,wang2025vggt}.
The operator $\mathcal Q_{.98}$ trims the upper tail by averaging only values below the 98th percentile, independently for the confidence-weighted and unweighted terms.
During training, frames are sampled in random order so that any labelled observation can serve as the reference.
Implementation and training details are given in \cref{sec:supp_training}.

\subsection{Dense 4D tracking and reconstruction}\label{sec:method_inference}

A single network forward pass predicts HPMs and depth jointly for all input frames.
Each HPM sample retains its source frame and pixel index when the maps are combined into a union (\cref{fig:method}).
For a query pixel $p$ in frame $t$, we search only the HPM samples belonging to the person mask in target frame $u$:
\begin{equation}\label{eq:nn}
\hat q(p;t\!\to\!u)=\argmin_{q\in\mathcal M_u}
\|\hat{\bX}_u(q)-\hat{\bX}_t(p)\|_2.
\end{equation}
We estimate a match as visible when the residual at $\hat q$ is below $\tau=0.005$.
This allows a small tolerance between predictions of the same surface point in normalised reference coordinates.
An estimate of the metric scale could instead support a tolerance specified in centimetres; we use a fixed normalised threshold.
The predicted depth lifts the matched pixel into the target frame's original camera coordinates, using the crop intrinsics and rotation:
\begin{equation}\label{eq:owntime}
\mathbf P_u(q)=\bR_{v,u}^{\top}\hat D_u(q)\bK_{v,u}^{-1}\tilde q,
\qquad
\hat{\mathbf T}_{t,p}(u)=\mathbf P_u\!\left(\hat q(p;t\!\to\!u)\right).
\end{equation}
Here $\tilde q=[q_x,q_y,1]^\top$, with $q_x,q_y$ denoting pixel-centre coordinates.
Repeating the lookup gives the query point's trajectory and visibility estimates in per-frame camera coordinates, with a shared normalised scale.
Each 3D position uses the depth predicted at the matched target pixel, so tracks and per-frame depth are consistent by construction.

The known crop warps and person masks align our predictions with the image pixels used by a scene reconstruction model.
Scene depth within the person masks sets the person-to-scene scale; scene camera poses then place our human geometry and tracks in the reconstructed world (\cref{sec:supp_composition}).
Our predicted depth and HPMs supply the human geometry and correspondences.
For internet video, we use VGGT-$\Omega$~\citep{wang2026vggtomega} for scene reconstruction and SAM~3~\citep{carion2025sam3} for person masks.
Benchmark-specific inter-person scaling is described in \cref{sec:supp_protocols_pstudio}.

\section{Experiments}\label{sec:experiments}

We evaluate the 4D representation through point tracking on real moving people and an amodal 3D localisation protocol on 4D-DRESS.
Crop-conditioning ablations and temporal analysis are reported in \cref{sec:supp_analysis}, with additional qualitative results in \cref{sec:supp_results}.

\subsection{Evaluation setup}\label{sec:exp_setup}

We train Dense4D using all 27,727 BEDLAM2~\citep{tesch2025bedlam2} training tracks, 2,200 of the 2,445 THuman2.1~\citep{yu2021function4d} scans and 15,710 MammaSyn~\citep{cuevas2025mamma} person tracks.
Every tenth THuman2.1 scan is held out; the training budget and sampling schedule are given in \cref{sec:supp_training}.
We use BEDLAM2's official test split for in-domain ablations, while our main evaluations assess transfer to real footage.

\subsection{4D point tracking on Panoptic Studio Humans}\label{sec:exp_pstudio}

\paragraph{Protocol.}
We evaluate the human-query subset of TAPVid-3D Panoptic Studio~\citep{koppula2024tapvid3d,joo2019panoptic}, which we refer to as \emph{Panoptic Studio Humans}.
We retain all 156 clips of 150 frames, evaluating the same set of queries on people across methods.
We report average Jaccard (AJ), average percentage of points within the depth-adaptive thresholds (APD), and occlusion accuracy (OA), using the official evaluator.
VGGT-$\Omega$ depth sets the relative scale of people; our model supplies their geometry and correspondences.
We adapt the outputs of V-DPM, 4RC and Any4D to the benchmark camera intrinsics without using ground-truth 3D.
Further protocol details are given in \cref{sec:supp_protocols_pstudio}.

\begin{table}[t]
\centering
\begin{minipage}[c]{0.36\textwidth}
\phantomcaption\label{tab:pstudio}
\centering\footnotesize
\setlength{\tabcolsep}{3.5pt}
\renewcommand{\arraystretch}{1.17}
\begin{tabular}{@{}lrrr@{}}
\toprule
Method & AJ$\uparrow$ & APD$\uparrow$ & OA$\uparrow$\\
\midrule
\method & \best{38.4} & \best{50.3} & \best{89.6}\\
\midrule
TAPIP3D & 17.0 & 27.1 & 85.2\\
V-DPM\textsuperscript{\textdagger} & 16.9 & 27.6 & 76.7\\
SpatialTrackerV2 & 15.5 & 24.4 & 78.7\\
4RC\textsuperscript{\textdagger} & 14.4 & 24.1 & 76.6\\
Shape of Motion & 12.5 & 21.3 & 80.7\\
Any4D\textsuperscript{\textdagger} & 7.3 & 13.6 & 76.6\\
\bottomrule
\end{tabular}
\end{minipage}\hfill
\begin{minipage}[c]{0.625\textwidth}
\captionsetup{type=figure}
\phantomcaption\label{fig:tracking}
\centering
\begin{tikzpicture}
\node[anchor=south west,inner sep=0] (pstudio) at (0,0)
  {\includegraphics[width=\linewidth]{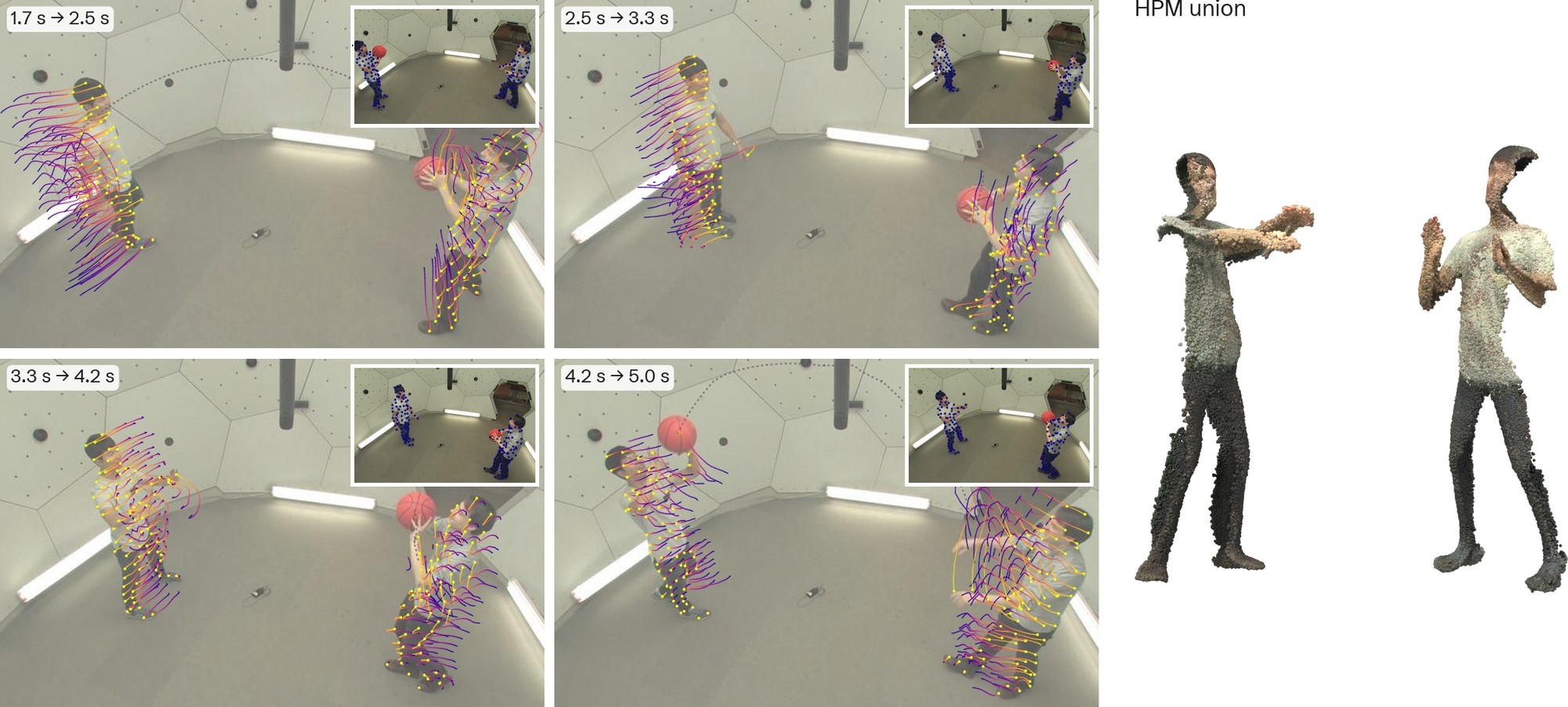}};
\fill[white] (pstudio.north east) rectangle
  ($(pstudio.north west)!0.70!(pstudio.north east)+(0,-12pt)$);
\node[anchor=north,inner sep=0.5pt,font=\scriptsize]
  at ($(pstudio.north west)!0.85!(pstudio.north east)$) {HPM union};
\end{tikzpicture}
\end{minipage}
\caption*{Table~\ref{tab:pstudio} and Figure~\ref{fig:tracking}: \textbf{Panoptic Studio Humans.}
Left: tracking on human-body queries (\textdagger: adapted to benchmark intrinsics; Appendix~\ref{sec:supp_protocols_baselines}).
Centre: our tracks from 1.7 to 5.0\,s.
Far right: HPM unions combining person pixels from across the input sequence in each person's reference pose.
Despite the motion across frames, these points assemble into coherent bodies with aligned limbs and surface detail, making the quality of temporal alignment directly visible.}
\end{table}

\paragraph{Results.}
Our model obtains 38.4 AJ, 50.3 APD and 89.6 OA (\cref{tab:pstudio}).
The strongest evaluated baselines are TAPIP3D (17.0 AJ, 27.1 APD) and V-DPM (16.9 AJ, 27.6 APD).
Our model improves on the best evaluated baseline for AJ by 21.4 points and for APD by 22.7 points.
Dense4D also achieves the highest occlusion accuracy, at 89.6 OA.
\Cref{fig:tracking} shows our tracks and reference-time HPM unions, with further examples in \cref{fig:supp_pstudio_tracking_unions}.

\subsection{Amodal 3D localisation on 4D-DRESS}\label{sec:exp_4ddress}

Localising observations at a common reference time requires predictions for regions that may be hidden in the reference image.
We introduce an \emph{amodal 3D localisation} evaluation on 4D-DRESS~\citep{wang2024ddress}, whose synchronised images and scans let us test whether points observed across time agree with the person's full reference-time geometry.
Because the scan covers regions hidden in the reference image, observations revealed later in the video can also be evaluated in the reference pose.
Small distances indicate close agreement with measured geometry, providing a stricter test than coarse localisation near the body.

\begin{figure}[!htb]
\centering
\input{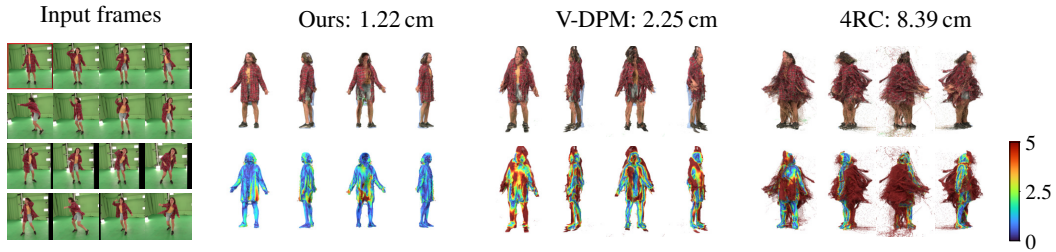}
\caption{\textbf{Reference-time reconstruction from sixteen observations on 4D-DRESS.}
Left: input frames, with the reference boxed.
For each method, four viewpoints show the union coloured by source appearance (top) and distance to the scan from 0 to 5\,cm (bottom).
The measured scan is overlaid in blue-grey in the top row.}
\label{fig:ddress}
\end{figure}

We evaluate 128 distinct dynamic takes, with one sixteen-frame window per take beginning at its first frame.
All methods receive the same person crops from one camera, with the first frame as the reference.
HPMs directly predict the reference-time positions of all foreground pixels; V-DPM~\citep{sucar2026vdpm} and 4RC~\citep{luo2026fourrc} decode each frame at that same reference time.
We combine the sixteen point maps into one cloud and report the mean one-sided point-to-scan distance, using the full reference scan.
The distance is one-sided because the input views need not cover the full reference scan.
Each method discards its own least confident 5\,\% of points across the whole union.
This one-sided point-to-scan distance measures geometric agreement across time, though proximity to the scan alone cannot establish exact point identity. We assess temporal matching through point tracking on Panoptic Studio Humans (\cref{sec:exp_pstudio}) and discuss the limits of these evaluations in \cref{sec:limitations}.

\begin{figure}[!htb]
\centering
\begin{minipage}[c]{0.29\textwidth}
\captionsetup{type=table}
\phantomcaption\label{tab:ddress}
\centering\small
\renewcommand{\arraystretch}{1.15}
\setlength{\tabcolsep}{1.5pt}
\begin{tabular*}{\linewidth}{@{\extracolsep{\fill}}lrrr@{}}
\toprule
Method & \shortstack{Mean\\(cm)$\downarrow$} & \shortstack{$>2$\,cm\\(\%)$\downarrow$} & \shortstack{$>5$\,cm\\(\%)$\downarrow$}\\
\midrule
V-DPM & 2.31 & 41.5 & 9.1\\
4RC & 4.84 & 53.4 & 20.7\\
\method & \best{0.86} & \best{6.8} & \best{0.1}\\
\bottomrule
\end{tabular*}
\end{minipage}\hfill
\begin{minipage}[c]{0.69\textwidth}
\captionsetup{type=figure}
\phantomcaption\label{fig:ddress_curves}
\includegraphics[width=\linewidth]{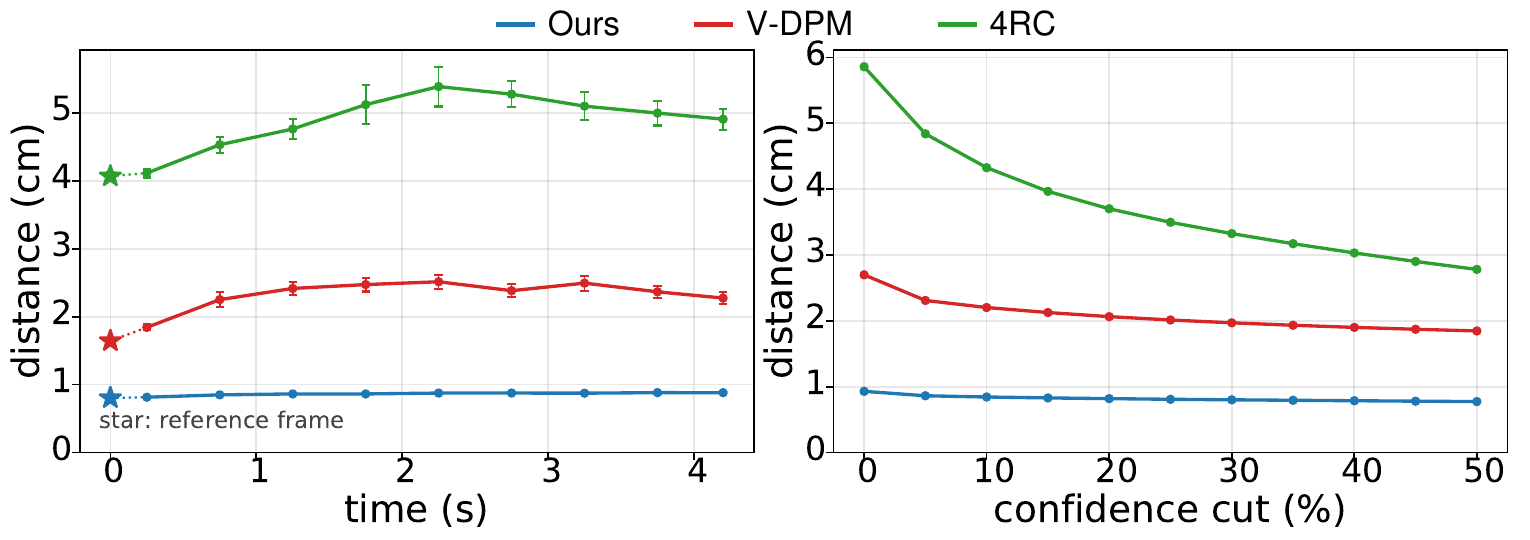}
\end{minipage}
\caption*{Table~\ref{tab:ddress} and Figure~\ref{fig:ddress_curves}: \textbf{Amodal 3D localisation on 4D-DRESS.}
Left: mean distance to scan and percentages of points more than 2 and 5\,cm from the surface.
Centre: error by time; right: error after dropping each method's least confident points.
Bars show standard errors; the table and temporal plot use a union-wide 5\,\% cut.}
\end{figure}

\paragraph{Results.}
Our union lies less than 1\,cm from the measured surface on average: 0.86\,cm, compared with 2.31\,cm for V-DPM and 4.84\,cm for 4RC (\cref{tab:ddress}).
Even with identical person-centred crops and little global translation, Dense4D achieves substantially more accurate surface localisation than the evaluated dynamic scene models.
Points from non-reference frames are nearly as accurate as those from the reference frame (0.87 versus 0.80\,cm), and only 6.8\,\% lie more than 2\,cm from the surface, compared with 41.5\,\% and 53.4\,\% for the baselines.
For both baselines, reference-frame predictions have lower error than those from other frames: 1.64 versus 2.36\,cm for V-DPM and 4.07 versus 4.89\,cm for 4RC.

\Cref{fig:ddress_curves} shows how error varies with time and confidence.
Our error stays within approximately 0.1\,cm of the reference-frame value across the plotted temporal range.
V-DPM's error rises by about 0.8\,cm within the first second and then levels off, while 4RC's grows gradually, by about 1.3\,cm over the first two seconds.
Increasing the confidence cut from 0 to 50\,\% reduces 4RC's error from 5.86 to 2.78\,cm and V-DPM's from 2.70 to 1.85\,cm, while ours changes from 0.93 to 0.77\,cm.
Confidence filtering therefore reduces the baselines' errors more strongly, while our unfiltered predictions already lie within 1\,cm of the surface on average.
\Cref{fig:ddress} shows one test case for the three methods, with further examples in \cref{fig:supp_ddress_unions}.

\subsection{Human surface consistency in the wild}\label{sec:exp_wild}

\begin{figure}[!htb]
\centering
\includegraphics[width=\textwidth]{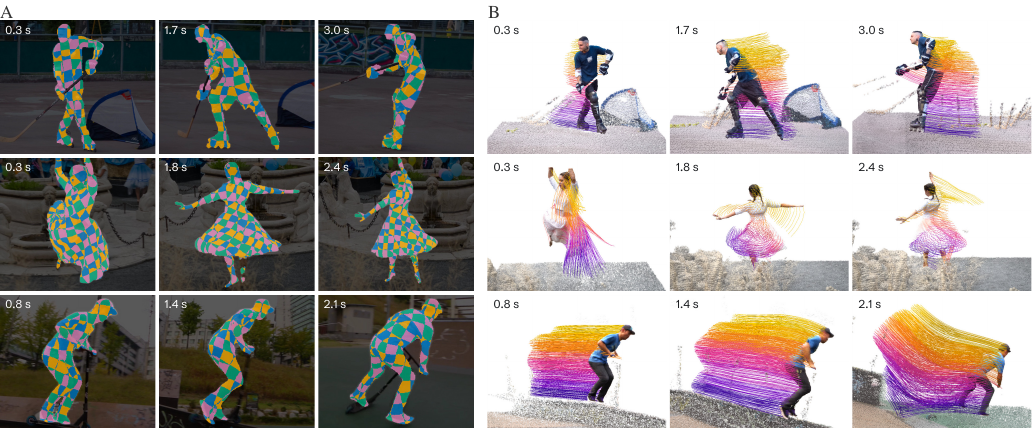}
\caption{\textbf{Dense correspondences and 4D tracks.}
Each row follows a person through three instants, with matching times shown in A and B.
(A) The same reference-space 3D chequerboard colours every image, making it possible to follow surface regions through body rotation and garment deformation.
(B) The reconstructed person and coloured 3D trajectories, integrated with VGGT-$\Omega$'s scene points.
Together, the panels connect correspondence on the image with the motion of human surface points in 3D.}
\label{fig:qualitative}
\end{figure}

\begin{figure}[!htb]
\centering
\includegraphics[width=\textwidth]{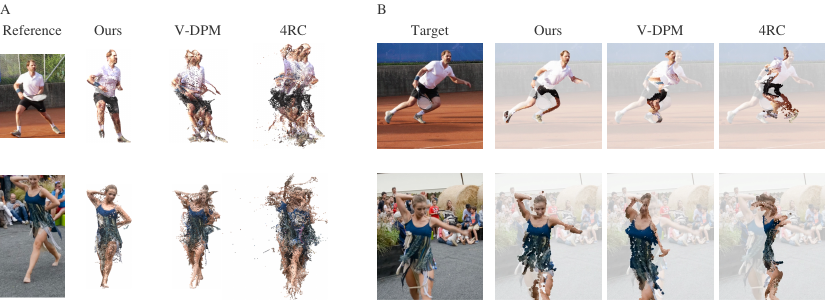}
\caption{\textbf{Human surface consistency on DAVIS.}
(A) Reference image and unions of person points from multiple frames, all expressed at frame zero.
Notice the distinct body outline preserved by Dense4D, compared with the overlapping limbs and dispersed points in the baseline unions.
(B) Frame-zero points transferred to the target image, with their original colours retained against a dimmed background.
These colours allow direct inspection of alignment between corresponding body regions.
Dense4D anchors transfers to pixels within the target person mask and uses their predicted depth, ensuring consistency with its reconstructed geometry by construction.}
\label{fig:surface_comparison}
\end{figure}

On DAVIS~\citep{pont2017davis}, Dense4D's transferred points follow the moving person more faithfully than those of V-DPM and 4RC, whose transfers exhibit displaced or repeated body parts.
\Cref{fig:surface_comparison} summarises this visual advantage with matched forward-transfer overlays and reference-time unions, where our predictions form a more coherent human surface.
Additional sequences illustrate body rotation and clothing deformation (\cref{fig:teaser,fig:qualitative,fig:supp_davis_correspondences}).
The same reference-space matching supports tracking over long intervals: Dense4D processes 1,287 person crops from an 86\,s video in one forward pass (\cref{fig:long_row}).

The comparisons distinguish plausible reconstruction in each frame from consistent transfer of the same surface points across time.
The additional views in \cref{fig:davis_forward} show this directly: the evaluated dynamic scene systems can reconstruct recognisable bodies yet place transferred points away from their own surfaces.

\section{Discussion and conclusion}\label{sec:limitations}\label{sec:conclusion}

Dense4D recovers dense 4D correspondence across the body, clothing and hair without a body template.
Its Human Point Maps assemble observations from different frames into coherent human geometry in the reference pose, with aligned body parts and fine surface detail.
Dense4D separates correspondence from camera-pose and scene-motion estimation while keeping tracks consistent with its predicted human geometry by construction.
Compared with the baselines, Dense4D produces more precise trajectories and a better fit to measured human geometry, with videos showing robust tracking through articulation and clothing deformation.
Dense4D can also be integrated with scene point clouds simply through shared image pixels.

\paragraph{Limitations.}
Self-contact can make reference-space correspondence ambiguous when distinct surface points coincide.
Selecting a reference frame without self-contact, when available, is a relatively straightforward mitigation that we do not pursue here.
A broader challenge is evaluating correspondence on real people: 4D-DRESS measures proximity to scans rather than point identity, while Panoptic Studio Humans relies on reconstruction-derived trajectory labels that may inherit reconstruction errors.
Synthetic data provides exact correspondence, but domain differences limit its relevance to real images.
Evaluation of scene integration remains open as well: our visualisations demonstrate effective integration with scene point clouds, but its accuracy has not yet been measured quantitatively.

\paragraph{Future work.}
Temporal and multi-view consistency losses on large corpora of real video could improve correspondence and reduce reliance on synthetic supervision.
Developing real-data benchmarks with independently measured surface-point identities would strengthen evaluation beyond surface proximity and reconstruction-derived tracks.
The alignment provided by Dense4D could support learning appearance models based on Gaussian splatting~\citep{kerbl2023gaussians} from humans in motion, while its accurate tracks could control human motion in generated video~\citep{geng2025motionprompting}.
Dense4D provides a foundation for augmenting scene understanding systems with more faithful human geometry, dense correspondence and robust tracking.
Building on this foundation, future work could develop richer models of people and their surroundings that remain coherent through prolonged motion and interaction.

\label{page:main_end}

\section*{Use of large language models}
LLM-based tools assisted with implementing components of the system, running and organising experiments, and writing scripts for visualisations using 3D graphics tools.
They also assisted with editing and polishing the manuscript and occasionally supported literature and dataset searches.
We conceived the research, developed the representation and methodology, designed the experiments and interpreted the results.
We directed the computational work, reviewed the resulting code and manuscript, and take full responsibility for the paper's claims and conclusions.

\section*{Ethics statement}
This work uses existing research datasets and publicly available video footage, including clips from \citet{pexels}, in accordance with their respective terms of use.
No new recordings of people were collected.
Training uses BEDLAM2, THuman2.1 and MammaSyn; evaluation includes existing benchmarks of real videos and 3D scans.
Dense4D estimates human geometry and correspondence without predicting biometric identity.
Applications should respect the consent and privacy of the people depicted.

Our experiments do not assess demographic differences in performance or suitability for consequential decisions about individuals.
The underlying datasets and pretrained models remain subject to their respective terms of use.

\section*{Reproducibility statement}
The representation, training targets and inference procedure are defined in \cref{sec:method}.
The appendix details data preparation (\cref{sec:supp_data}), implementation and training (\cref{sec:supp_method}), and evaluation protocols and baseline settings (\cref{sec:supp_protocols}).
We will publicly release our inference code and evaluation pipelines, together with code to reproduce the image and video visualisations.

\bibliography{main}
\bibliographystyle{iclr2027_conference}

\clearpage
\appendix
\crefalias{section}{appendix}
\crefalias{subsection}{appendix}
\section{Data and surface supervision}\label{sec:supp_data}

\begin{figure}[!htbp]
\centering
\includegraphics[width=\textwidth]{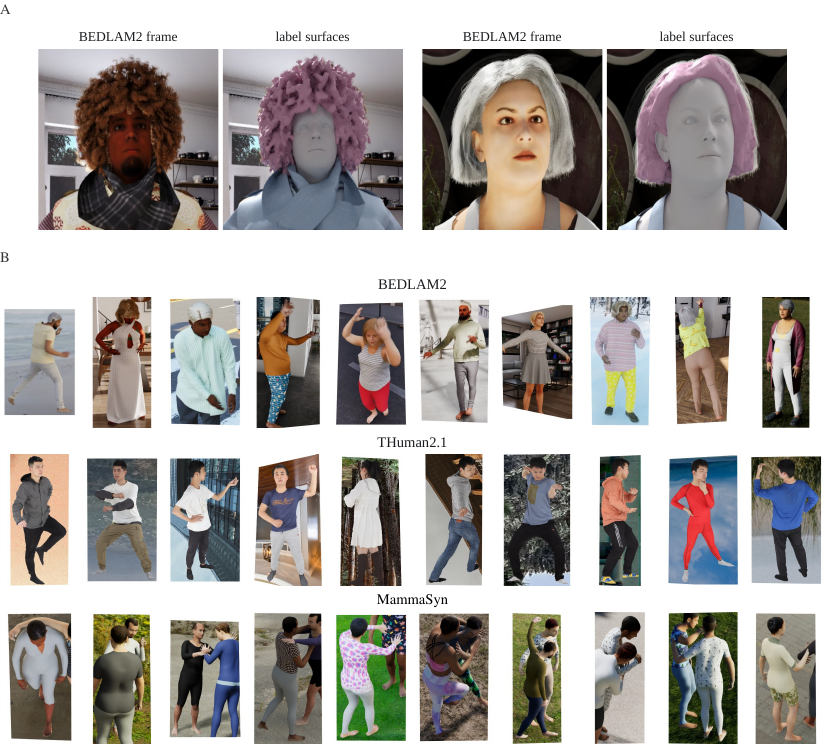}
\caption{\textbf{Training surfaces and samples.}
(A) BEDLAM2 images and the body, garment and hair surfaces used for supervision.
(B) Training crops from BEDLAM2, THuman2.1 and MammaSyn, each from a different clip or scan.}
\label{fig:training_data}
\end{figure}

All three training sources use the surface-label construction in \cref{sec:method_gt}: triangle indices and barycentric coordinates provide both HPM and depth targets.
We rasterise all people together with a shared depth buffer to account for inter-person occlusion.
\Cref{fig:training_data} illustrates the supervision surfaces and input crops.

\subsection{BEDLAM2}\label{sec:supp_bedlam_data}
Holding out every tenth BEDLAM2~\citep{tesch2025bedlam2} track leaves 27,727 training tracks (1,250,307 labelled frames) and 3,081 development tracks, with 1,454 subjects across the two subsets.
The official test split contains 7,157 tracks and 181,263 labelled frames from 1,223 subjects.

Body and camera labels are available every fifth frame of the 30\,fps video, giving supervision at 6\,Hz.
The surfaces comprise neutral SMPL-X bodies~\citep{pavlakos2019smplx}, garments with fixed topology across time, and hair envelopes derived from strand grooms.

\begin{figure}[!htbp]
\centering
\includegraphics[width=0.88\textwidth]{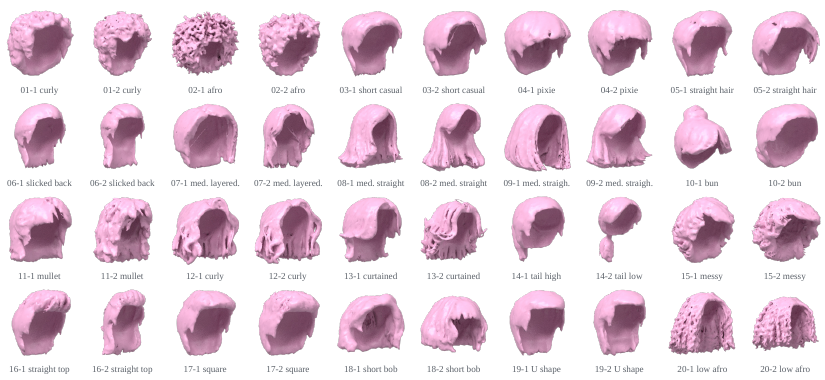}
\caption{\textbf{Hair envelope meshes.}
Envelopes derived from BEDLAM2's strand grooms, shown from a common viewpoint with consistent lighting.
Labels identify the source grooms.}
\label{fig:supp_hairmeshes}
\end{figure}

For each groom, we extract the 0.3 level set of a strand-density field, offset it 3\,mm inward and decimate it to 25\,\% of its faces~\citep{lorensen1987marching,garland1997surface}.
We observe no independently simulated hair motion in the BEDLAM2 renders.
The envelope vertices are bound to head triangles and transported with the posed head.
\Cref{fig:supp_hairmeshes} shows the envelopes used in \cref{fig:training_data}(A).
Compared with the renderer's hair mattes, the envelopes achieve approximately 92\,\% precision and 84\,\% recall.

\begin{figure}[t]
\centering
\includegraphics[width=.9\textwidth]{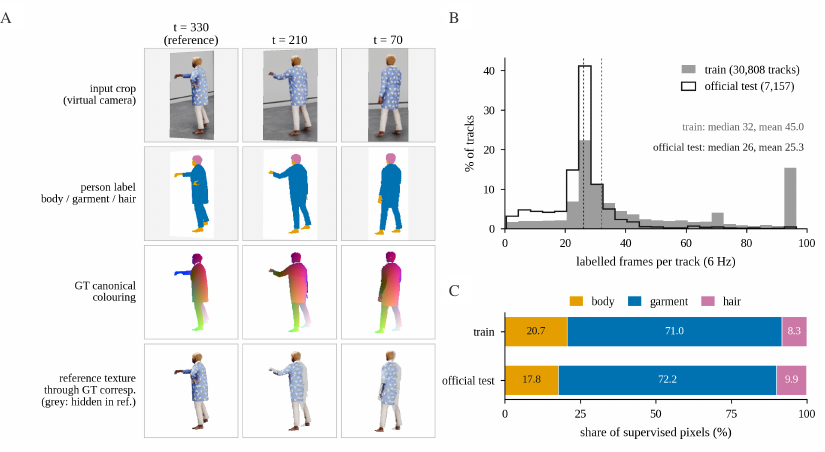}
\caption{\textbf{Surface supervision for 4D correspondence.}
A human sequence with body, garment and hair labels, reference-coordinate colours and texture transfer.
The plots summarise track length and labelled-pixel composition; the histogram's training pool includes the development tracks.}
\label{fig:supp_gt}\label{fig:data}
\end{figure}

Body, garment and hair pixels account for 21, 71 and 8\,\% of training labels and 18, 72 and 10\,\% of test labels, estimated from 1,541 training and 1,431 test tracks (\cref{fig:supp_gt}).

\subsection{THuman2.1}\label{sec:supp_thuman_data}
We train on 2,200 of the 2,445 textured THuman2.1~\citep{yu2021function4d} scans, holding out every tenth scan.
Each scan is rendered from 50 cameras on a jittered orbit: elevation $[-10^\circ,25^\circ]$, log-uniform vertical field of view $[25^\circ,90^\circ]$, off-centre aim and projected extent of 55 to 85\,\% of image height.
Images are $1280\times720$ with panorama backgrounds sampled by camera ray.
RGB and labels share the same rasterisation.

\subsection{MammaSyn}\label{sec:supp_mammasyn_data}
We train on 15,710 person tracks in 7,856 views from MammaSyn, MAMMA's synthetic data family~\citep{cuevas2025mamma}.
We retain cameras pitched downwards by at least $15^\circ$ and every fourth frame.
MammaSyn complements BEDLAM2's predominantly eye-level views and limited interactions with elevated views and interacting people, but its photorealistic SMPL-X renderings represent clothing only as texture, without separate garment or hair meshes.

\section{Implementation and training}\label{sec:supp_method}

\subsection{Architecture and virtual cameras}
A $4\to1024\to1024$ MLP with GELU activation embeds the crop geometry.
The ViT image encoder uses DINOv3-L~\citep{simeoni2025dinov3} with patch size 16.
Each $512\times512$ crop supplies 1,024 patch tokens of width 1,024, which are joined by one camera token, 16 register tokens and the crop token.
Separate learned camera and register tokens identify the first frame as the reference; the remaining frames share a second set.
Camera tokens are internal features, with no camera-prediction head or camera-regression loss.
There are 24 pairs of frame and inter-frame attention blocks.
Five inter-frame blocks (indices 2, 6, 9, 14 and 20, zero-indexed) update only the camera, register and crop tokens; patch tokens pass through these steps unchanged.
The remaining 19 inter-frame blocks attend across all tokens.

Both dense heads use features from blocks 4, 11, 17 and 23 (zero-indexed), projected to resolutions $128^2$, $64^2$, $32^2$ and $16^2$, then fused by a 256-channel RefineNet decoder~\citep{lin2017refinenet} and pixel-shuffled to $512^2$.
HPM coordinates have no output activation.
The heads run in fp32 and the backbone in bf16; each head has 32.3\,M parameters.

The virtual-camera warp follows \cref{eq:cond}, with supervision extrinsics $\mathbf E_{v,t}=\bR_{v,t}\mathbf E_t$, where $\mathbf E_t$ is the original world-to-camera transform.
Training boxes have 12\,px padding; test boxes are smoothed over seven frames and expanded by 12\,\% on each side.
Images are warped bilinearly and labels/masks by nearest neighbour.

\subsection{Integration with dense scene reconstruction}\label{sec:supp_composition}
To place our human geometry in the scene, we use VGGT-$\Omega$'s full-frame depth $D_t^\Omega$ and world-to-camera pose $(\bR_t^\Omega,\bt_t^\Omega)$.
The crop warp and person mask pair scene-depth samples with the real-camera depths of our predicted points $\mathbf P_t$ from \cref{eq:owntime}.
These depth pairs determine a person-to-scene scale $s_{w,t}$, giving
\begin{equation}\label{eq:composite}
\bX_t^{\rm world}=\bR_t^{\Omega\top}(s_{w,t}\mathbf P_t-\bt_t^\Omega).
\end{equation}
Scene depth sets the scale only; our predicted depth supplies the human geometry.
Background points come from the scene reconstruction outside the person masks.
For Panoptic Studio Humans we fit one scale per person over the clip and score tracks in the benchmark cameras (\cref{sec:supp_protocols_pstudio}).
For the moving-camera visualisations, $s_{w,t}$ is the per-frame median ratio of VGGT-$\Omega$ depth to our depth over the eroded person mask.
\Cref{tab:supp_cost} reports inference time and peak GPU memory for the full pipeline.

\begin{table}[t]
\centering
\footnotesize
\setlength{\tabcolsep}{4.5pt}
\caption{Inference time and memory of the full pipeline on one GH200, for a $1280\times720$ internet video, with
each stage at its own input size. SAM~3 tracks the person on the full frames; VGGT-$\Omega$ gives the cameras and
scene depth from the full frames at $512\times288$; Dense4D runs on $512\times512$ person crops, which carry
1.8$\times$ the tokens of VGGT-$\Omega$'s frames, so its forward pass is the longer one although the two networks use
the same backbone architecture. ``Crops, read-out'' is CPU time for building the crops and writing the outputs. All models stay
loaded in one process (loading, 27\,s, is not included); peak memory is the GPU allocator maximum, including
11.4\,GB of model weights.}
\label{tab:supp_cost}
\begin{tabular}{@{}rrrrrrr@{}}
\toprule
 & SAM~3 & VGGT-$\Omega$ & \multicolumn{2}{c}{Dense4D, $512\times512$ crops} & & \\
\cmidrule(lr){4-5}
Frames & $1280\times720$ & $512\times288$ & crops, read-out & forward & Total & Peak memory \\
\midrule
100 & 10.6\,s & 1.8\,s & 0.8\,s & 4.7\,s & 17.9\,s & 19.5\,GB \\
150 & 18.2\,s & 3.3\,s & 1.4\,s & 10.0\,s & 32.9\,s & 21.9\,GB \\
300 & 33.0\,s & 11.0\,s & 2.6\,s & 36.6\,s & 83.2\,s & 29.1\,GB \\
\bottomrule
\end{tabular}
\end{table}

\subsection{Training schedule}\label{sec:supp_training}
Dense4D is trained for a cumulative 686,000 steps, taking approximately 490 hours on four GH200 GPUs.
We report the final checkpoint, without weight averaging.
AdamW~\citep{loshchilov2019decoupled} uses weight decay 0.05 and peak learning rate $2\times10^{-5}$, with linear warm-up and cosine decay.
Gradient norms are clipped to 1.0 separately for the backbone and each head; activation checkpointing covers attention-block pairs.

The final training configuration samples BEDLAM2, THuman2.1 and MammaSyn with probabilities of approximately 65\,\%, 22\,\% and 13\,\%, respectively.
For each person, the number of input frames is sampled from $S\in\{2,3,4,6,8,12,24\}$ with weights $1,1,1,1,2,3,3$.
The batch contains 96 images per step, consistently packed as 24 images per GPU; the number of person sequences varies with $S$.
Frames are sampled without replacement where the track length permits.

Photometric augmentation uses colour jitter with probability 0.9 and greyscale with probability 0.05; jitter is shared across frames with probability 0.7.
Brightness, contrast and saturation factors lie in $[0.5,1.5]$, with hue in $[-0.1,0.1]$.

\section{Evaluation protocols}\label{sec:supp_protocols}

\subsection{Panoptic Studio Humans}\label{sec:supp_protocols_pstudio}
We evaluate 40,779 human queries across all 156 clips of 150 frames, retaining the Dynamic-3DGS pseudo-labels~\citep{luiten2024dynamic3dgs}.
Queries on held objects or with invalid annotations are excluded consistently across methods.
The official TAPVid-3D evaluator uses benchmark intrinsics, depth-adaptive thresholds and one median rescaling per clip; scores are averaged over clips.
The dataset contains 49 two-person and 107 single-person clips.

SAM~3~\citep{carion2025sam3} supplies person masks, and the frame with the largest mask is selected as each person's reference.
We fit one scale per person across the clip using pixel-aligned VGGT-$\Omega$ and predicted human depths on confident mask interiors.
The fit selects the scale supported by the most depth ratios within 5\,\%, then refines it using the median inlier ratio.
VGGT-$\Omega$ sets the scale only; our predicted depth supplies the track geometry, without temporal smoothing.

A match is visible when its reference-space residual is below $\tau=0.005$.
Undoing the normalisation in \cref{eq:normalisation} gives a tolerance of $s\tau$: for example, $s=2$\,m gives 1\,cm.

\subsection{Baseline implementations}\label{sec:supp_protocols_baselines}
All baselines use the same human queries and evaluator.

\paragraph{4RC.}
4RC~\citep{luo2026fourrc} encodes all 150 frames jointly and decodes each query at each target time.
We project its points with its own intrinsics, then lift the resulting 2D tracks at unchanged depth using benchmark intrinsics.
This gives 14.4 / 24.1 AJ / APD, versus 4.1 / 7.4 for its original camera-space points; it has no occlusion output.

\paragraph{V-DPM.}
V-DPM~\citep{sucar2026vdpm} processes pairs drawn from 14 temporal chunks of at most 11 frames, with frame zero as reference; we align scales on shared reference pixels. Confidence $\geq1.5$ marks visibility.
We reproject each query's predicted position at its query frame using benchmark intrinsics, giving $(u,v)$; $(u_q,v_q)$ are the supplied query pixels.
Without using ground-truth 3D, we fit $u-c_x\approx k_u(u_q-c_x)+b_u$ and $v-c_y\approx k_v(v_q-c_y)+b_v$ by least squares per clip, then set $k=(k_u+k_v)/2$.
We correct points as $x'=(x-z\,b_u/f_x)/k$, $y'=(y-z\,b_v/f_y)/k$, $z'=z$, where $f_x,f_y,c_x,c_y$ are the benchmark intrinsics.
AJ / APD rises from 8.9 / 14.9 uncorrected to 12.7 / 20.7 with focal correction alone and 16.9 / 27.6 with the fitted scale and image-plane offset correction (\cref{tab:pstudio}).

\paragraph{Other baselines.}
Any4D~\citep{karhade2025any4d} groups queries by reference frame and uses all targets; we fit its predicted rays to intrinsics and convert its points to the benchmark camera as for 4RC.
We evaluate SpatialTrackerV2~\citep{xiao2025spatialtrackerv2} using the authors' TAPVid-3D settings, with benchmark intrinsics and a static camera.
To obtain tracks before the query frame, we run the tracker backwards through the video.
TAPIP3D~\citep{zhang2025tapip3d} uses its world-space MegaSaM~\citep{li2025megasam} variant, with the authors' annotations on 106 full-split clips and the released script on the other 50.
Shape of Motion~\citep{wang2025shapeofmotion} uses default optimisation, the fixed benchmark camera and person foreground masks; divergent boxes\_28 is scored as failed.

\subsection{4D-DRESS: amodal 3D localisation}\label{sec:supp_ddress}
Each of the 128 dynamic takes supplies sixteen frames from one camera, beginning at its first frame.
All methods receive identical crops and predict points at the first frame's reference time.

For each method, one similarity transform is robustly fitted using only reference-frame pixel correspondences to the scan, then applied unchanged to every frame's predictions.
Each method discards its least confident 5\,\% of points across the sixteen-frame union; reference-frame and temporal statistics use this same retained point set.
We report mean one-sided point-to-scan distance in centimetres and the fractions exceeding 2 and 5\,cm, averaging these statistics over takes.
The distance is one-sided because the observations do not cover the full scan; regions hidden in the reference image are included when observed in other frames.
Error bars are standard errors over takes.

\subsection{Evaluation using synthetic data}\label{sec:supp_protocols_bedlam}
The conditioning ablation uses 128 tracks selected at a fixed stride through BEDLAM2's sorted official test split, with four evenly spaced frames per track.
We measure Euclidean distance from each predicted point to its ground-truth reference-time position.
Known reference-camera geometry and each clip's ground-truth scale (\cref{eq:normalisation}) convert errors to centimetres before averaging, without fitted alignment or confidence filtering.
The temporal analysis uses the same error and scale convention (\cref{sec:supp_temporal}).

\section{Ablations and analysis}\label{sec:supp_analysis}

\subsection{Crop zoom and viewing direction}\label{sec:exp_cond}\label{sec:exp_generalist}
The crop embedding encodes both the centre ray and crop zoom (\cref{eq:cond}); we examine their roles separately.
Removing the entire embedding increases BEDLAM2 pointwise reconstruction error from 2.99 to 4.06\,cm and reduces Panoptic Studio Humans tracking from 38.4 to 30.4 AJ and from 50.3 to 41.4 APD (\cref{tab:ablation}).
The unconditioned variant has higher reconstruction error on 111 of 128 test tracks and lower tracking AJ on 150 of 156 clips.

Retaining crop zoom while fixing the three centre-ray channels to $(0,0,1)$ at inference gives equivalent reconstruction and tracking performance at the reported precision.
This is expected because the rotation-induced homography already reorients each crop towards the person.
The result suggests that crop zoom carries the useful conditioning information after this rectification.
The reported checkpoint retains the centre-ray inputs used during training, but their variation provides no measurable benefit in this experiment.

\begin{table}[!htbp]
\centering\small
\begin{tabular}{@{}lcccc@{}}
\toprule
Variant & \multicolumn{3}{c}{Panoptic Studio Humans} & BEDLAM2\\
\cmidrule(lr){2-4}
 & AJ$\uparrow$ & APD$\uparrow$ & OA$\uparrow$ & (cm)$\downarrow$\\
\midrule
Without crop conditioning & 30.4 & 41.4 & 89.3 & 4.06\\
Centre ray fixed to $(0,0,1)$ & \best{38.4} & \best{50.3} & \best{89.6} & \best{2.99}\\
Final model & \best{38.4} & \best{50.3} & \best{89.6} & \best{2.99}\\
\bottomrule
\end{tabular}
\caption{\textbf{Crop-geometry conditioning.}
The fixed-ray row modifies only the final model's inference input.
Evaluation follows \cref{tab:pstudio} for Panoptic Studio Humans and \cref{sec:supp_protocols_bedlam} for BEDLAM2.}
\label{tab:ablation}
\end{table}

\subsection{Temporal extent and input density}\label{sec:supp_temporal}
\begin{figure}[!htbp]
\centering
\includegraphics[width=\textwidth]{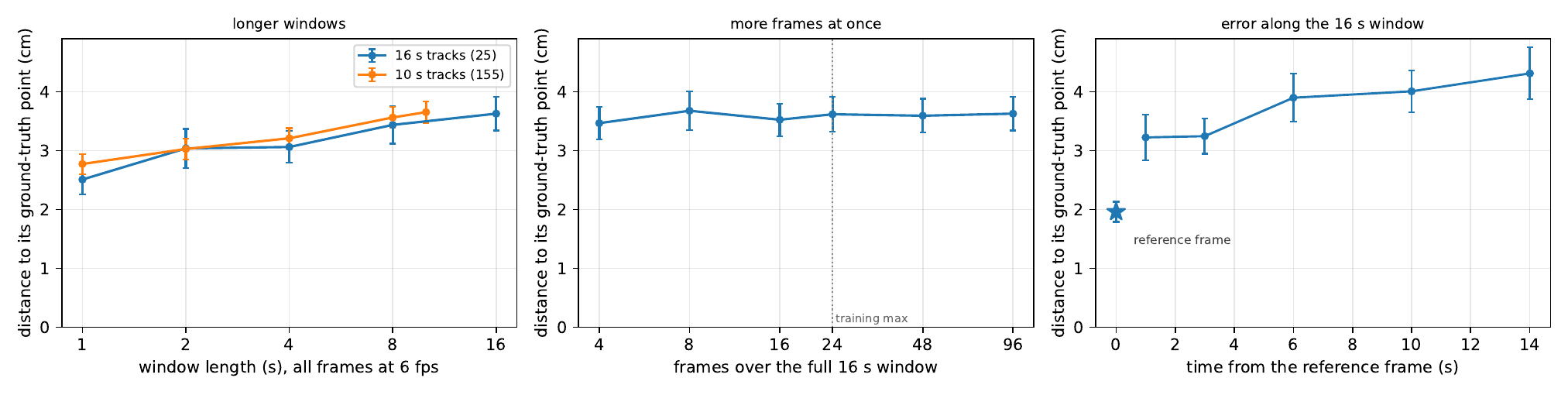}
\caption{\textbf{Temporal extent and input density on BEDLAM2.}
Pointwise reconstruction error in centimetres, with means and standard errors over tracks.
Left: windows sampled at 6\,fps on tracks lasting at least ten seconds (155 tracks) or sixteen seconds (25 tracks).
Middle: 4 to 96 frames over the same sixteen-second window.
Right: error by temporal offset within that window; the star marks the reference.}
\label{fig:duration}
\end{figure}

We use all official-test BEDLAM2 tracks lasting at least ten seconds (155 tracks, 139 subjects), including 25 that last sixteen seconds.
Windows start at the reference frame and span 1 to 16\,s, sampled at 6\,fps for up to 96 frames per forward pass.
Evaluation follows \cref{sec:supp_protocols_bedlam}; \cref{fig:duration} reports means and standard errors over tracks.
Over a fixed sixteen-second interval, varying the input from 4 to 96 frames gives reconstruction errors of 3.5 to 3.7\,cm, including frame counts above the usual training maximum of 24.
Extending the interval from one to sixteen seconds increases error from 2.5 to 3.6\,cm, while reference-frame error remains approximately 2.0\,cm.
Point-to-surface distance remains between 0.87 and 0.92\,cm, suggesting that the additional pointwise error primarily concerns localisation along the surface.

\section{Additional qualitative results}\label{sec:supp_results}\label{sec:supp_wild}

\par\medskip\noindent\begin{minipage}{\textwidth}
\subsection{Tracking over a long video}\label{sec:supp_long_video}
The video is sampled at 15\,fps, using $1280\times720$ images and $512\times512$ person crops.
VGGT-$\Omega$ supplies the scene reconstruction (\cref{sec:supp_composition}).

\par\medskip
\centering
\includegraphics[width=\textwidth]{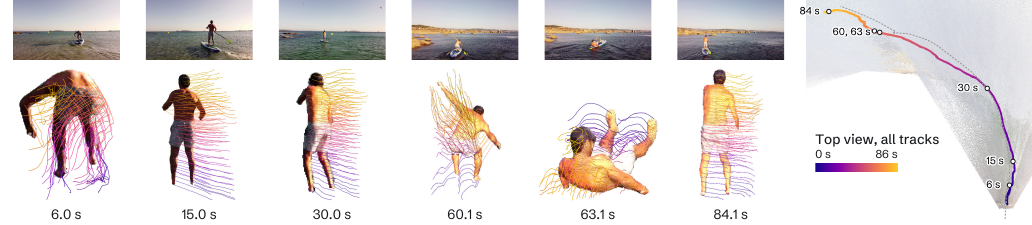}
\captionof{figure}{\textbf{Tracking over 86\,s.}
Left: input frames and one-second track segments coloured by body height.
Right: full-video tracks coloured by time in VGGT-$\Omega$'s scene.
The dashed line shows the camera path; circles mark the selected frames.}
\label{fig:long_row}
\end{minipage}

\clearpage
\subsection{Additional DAVIS visualisations}\label{sec:supp_davis_visuals}

For \cref{fig:surface_comparison}, we discard each method's least confident 5\,\% of points and use identical crops and display scales within each forward-transfer comparison.
Overlays are clipped at the target-image boundaries.

\noindent\begin{minipage}{\textwidth}
\centering
\includegraphics[width=\linewidth]{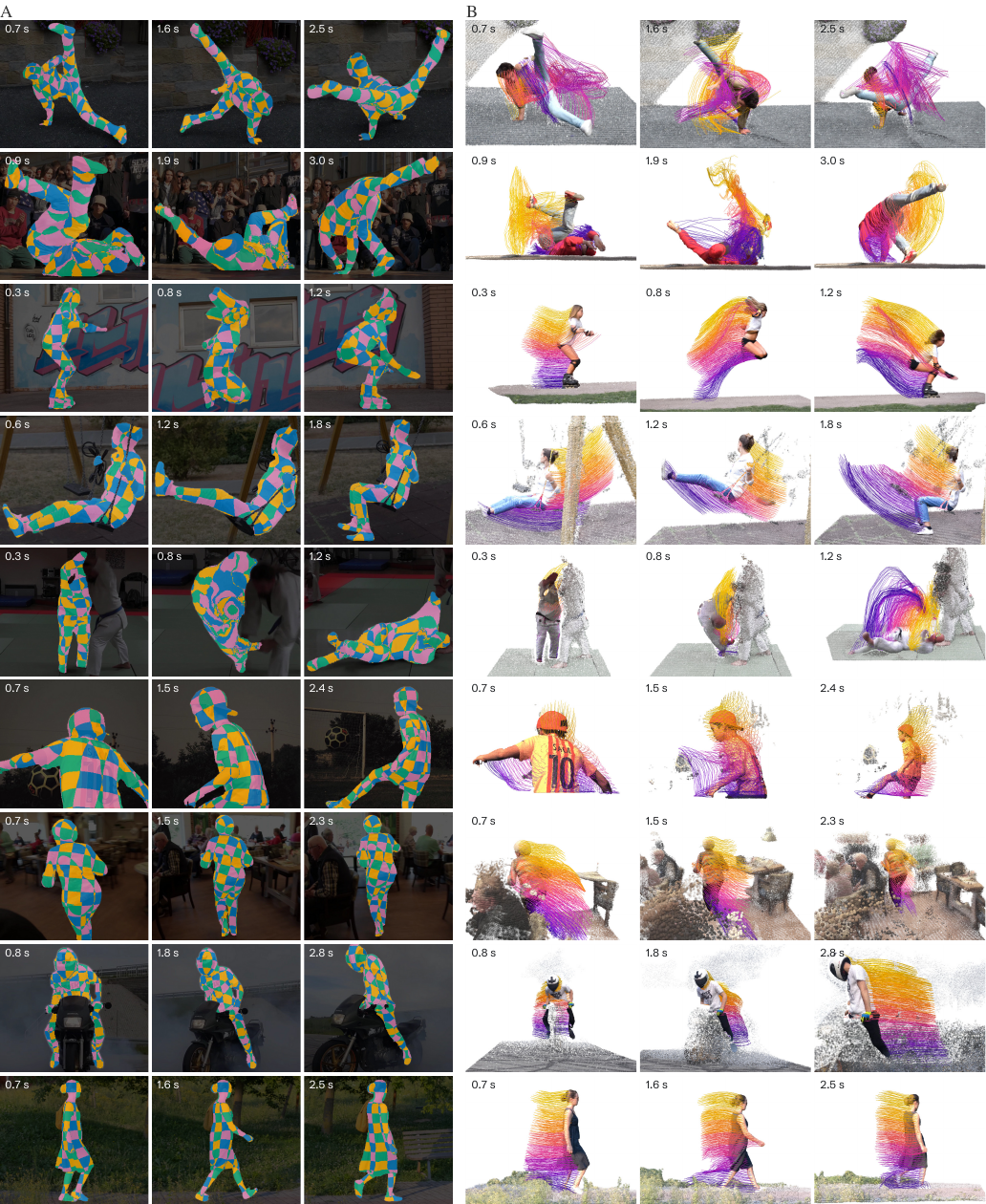}
\captionof{figure}{\textbf{Dense4D correspondences and 4D tracks on DAVIS.}
Each row shows three instants from one clip.
(A) A reference-space 3D chequerboard is transferred to the images through HPMs.
(B) Our 3D geometry and tracks in the reconstructed scene.}
\label{fig:supp_davis_correspondences}
\end{minipage}

\clearpage
\subsection{Forward-transfer consistency on DAVIS}\label{sec:supp_davis_forward}

\Cref{fig:davis_forward} compares frame-0 points transferred by Dense4D, V-DPM, 4RC and Any4D with each method's own target-time surface, extending the overlays in \cref{fig:surface_comparison}.
Several methods reconstruct recognisable human geometry at the target time, yet place transferred points away from that geometry.
This reveals inconsistency between their point-transfer and depth predictions, even when the target surface itself looks plausible.
Dense4D instead matches target-image pixels and lifts them with their predicted depth, so transferred points lie on its target surface by construction.
The height colours help assess alignment with the corresponding body regions; proximity to a method's own surface alone does not establish correspondence accuracy.

\par\medskip
\noindent\begin{minipage}{\textwidth}
\centering
\includegraphics[width=\linewidth]{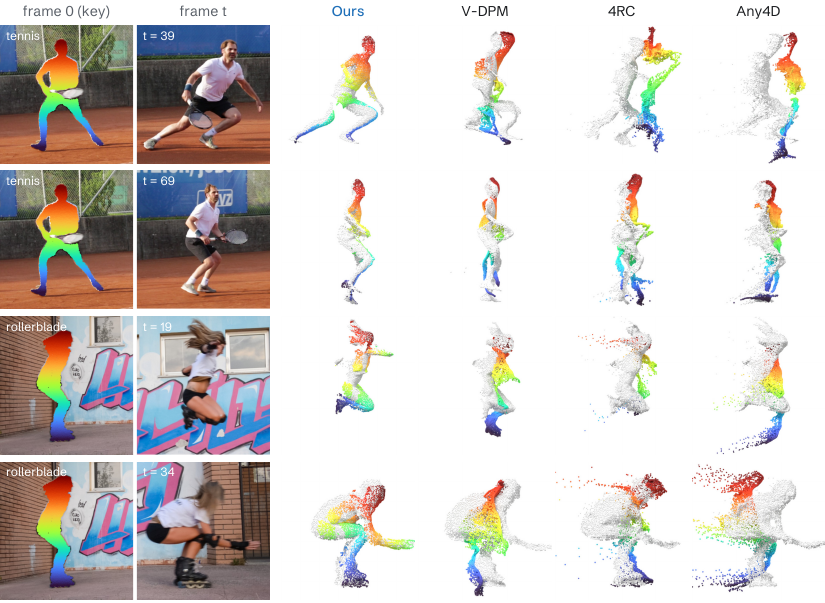}
\captionof{figure}{\textbf{Frame 0 carried forward on DAVIS.}
Left: frame~0 coloured by body height (head red, feet blue), and the target frame~$t$.
Right: each method's surface at $t$ (grey), overlaid with points from frame~0 transferred to $t$ and retaining their height colours.
Dense4D preserves the head-to-foot colour ordering through the displayed motions.
The baselines show scattered transferred points or displacement from their own target-time geometry.}
\label{fig:davis_forward}
\end{minipage}

\clearpage
\subsection{Additional Panoptic Studio Humans visualisations}\label{sec:supp_pstudio_visuals}

\noindent\begin{minipage}{\textwidth}
\centering
\includegraphics[width=\linewidth]{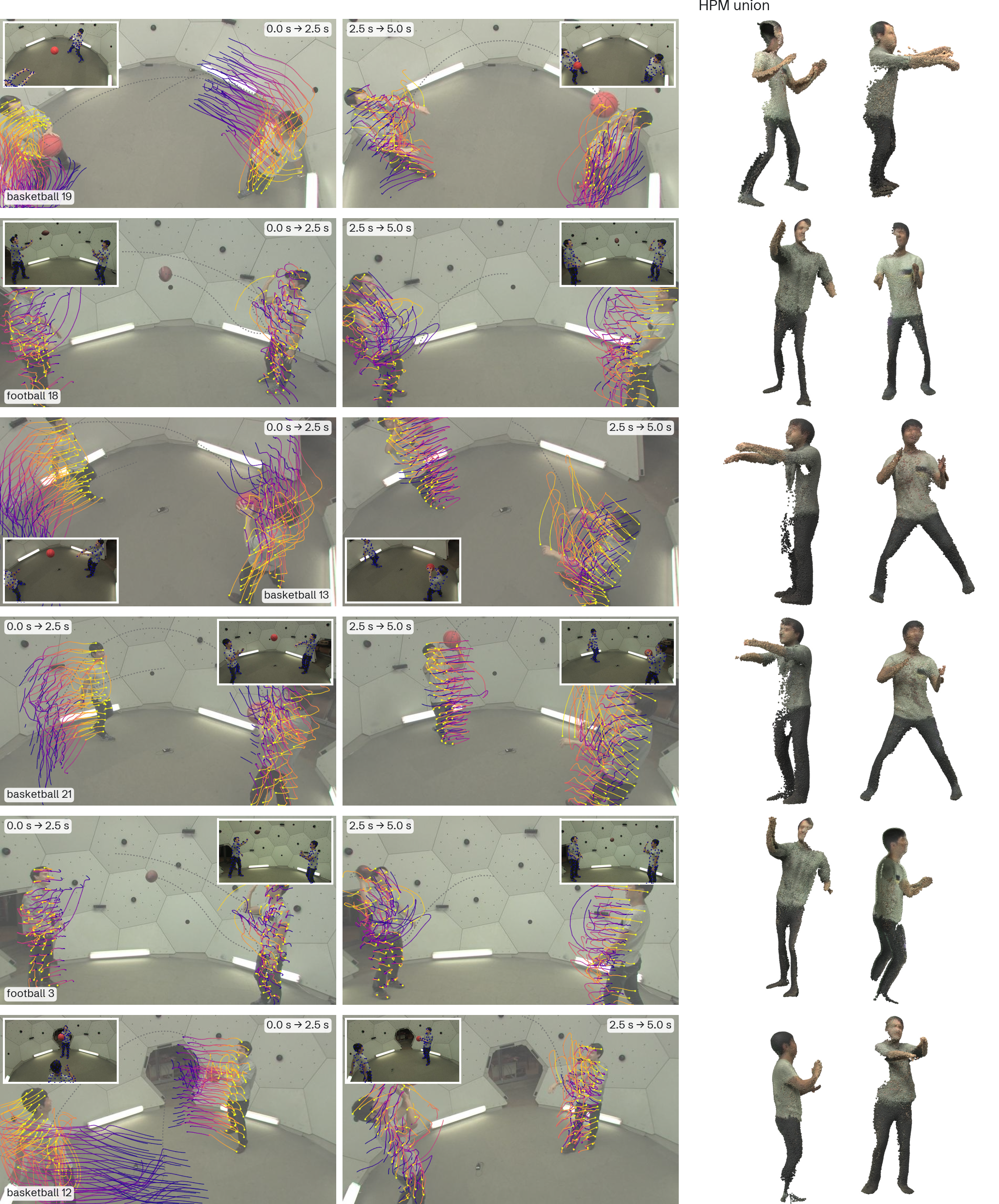}
\captionof{figure}{\textbf{Additional 4D tracking visualisations on Panoptic Studio Humans.}
Each row shows one clip, with tracks over 0.0--2.5\,s and 2.5--5.0\,s (left) and the two people's reference-time HPM unions (right).
Insets show image observations.}
\label{fig:supp_pstudio_tracking_unions}
\end{minipage}

\clearpage
\subsection{Additional 4D-DRESS visualisations}\label{sec:supp_ddress_visuals}

\noindent\begin{minipage}{\textwidth}
\centering
\includegraphics[width=\linewidth]{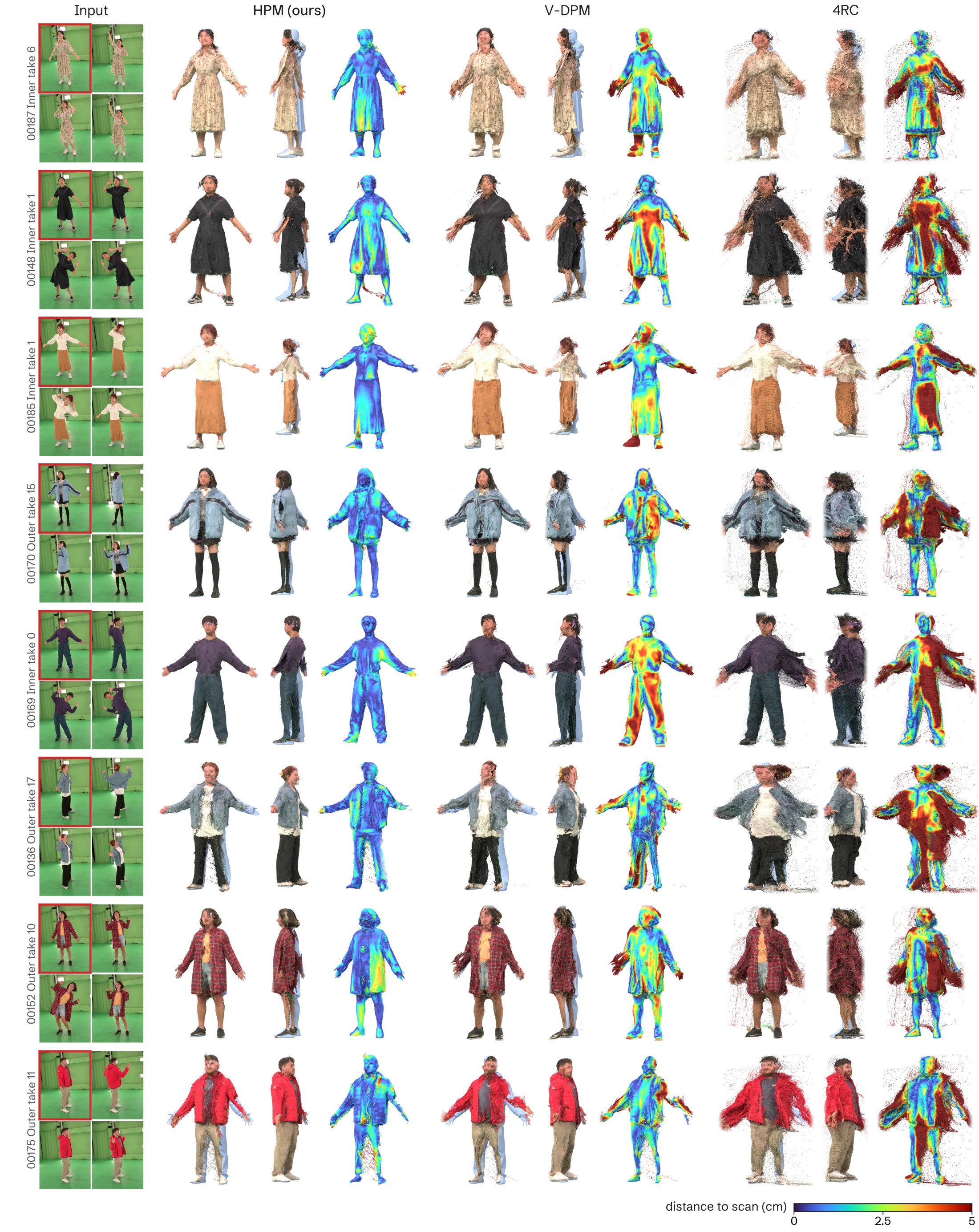}
\captionof{figure}{\textbf{Additional reference-time reconstructions on 4D-DRESS.}
Each row shows input images (reference boxed in red) and the unions predicted by Dense4D (HPM), V-DPM and 4RC.
For each method, two appearance views are followed by distance to the scan (0--5\,cm).
Each method discards its least confident 5\,\% of points.
The scan is overlaid in pale blue in the side view.}
\label{fig:supp_ddress_unions}
\end{minipage}

\end{document}